\pdfoutput=1

\def\year{2020}\relax
\documentclass[letterpaper]{article} 
\usepackage{aaai20}  
\usepackage{times}  
\usepackage{helvet} 
\usepackage{courier}  
\usepackage[hyphens]{url}  
\usepackage{graphicx} 
\def\UrlFont{\rm}  
\usepackage{graphicx}  
\usepackage{comment}
\usepackage{natbib}
\usepackage{color}

\title{HAVEN: A Unity-based Virtual Robot Environment to Showcase HRI-based Augmented Reality}

\author{Andre Cleaver\textsuperscript{\rm 1}, Darren Tang\textsuperscript{\rm 2}, Victoria Chen\textsuperscript{\rm 2}, Jivko Sinapov\textsuperscript{\rm 2} \\ 
\textsuperscript{\rm 1}Department of Mechanical Engineering\\ 
\textsuperscript{\rm 2}Department of Computer Science}

\begin{document}

\maketitle

\begin{abstract}
Due to the COVID-19 pandemic, conducting Human-Robot Interaction (HRI) studies in person is not permissible due to social distancing practices to limit the spread of the virus. Therefore, a virtual reality (VR) simulation with a virtual robot may offer an alternative to real-life HRI studies. Like a real intelligent robot, a virtual robot can utilize the same advanced algorithms to behave autonomously. This paper introduces HAVEN (HRI-based Augmentation in a Virtual robot Environment using uNity), a VR simulation that enables users to interact with a virtual robot. The goal of this system design is to enable researchers to conduct HRI Augmented Reality studies using a virtual robot without being in a real environment. This framework also introduces two common HRI experiment designs: a hallway passing scenario and human-robot team object retrieval scenario. Both reflect HAVEN's potential as a tool for future AR-based HRI studies.

\end{abstract}

\section{INTRODUCTION}

\begin{figure}[t!]
    \centering
    \includegraphics[width=0.35 \linewidth]{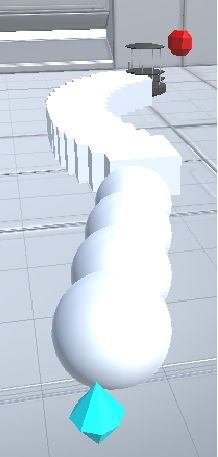} \caption{An AR visualization of the robot (Turtlebot2)'s path. White cubes represent the arc motion while the white spheres show straight motion. 
    }
\label{fig:path_projection}
\end{figure}

To establish common ground between robot and human, the human needs to understand the robot’s intention and behaviors. However, it is difficult for robots to interpret human intentions and effectively express their own intentions \citep{williams2019virtual}. To break the communication barrier, there are studies that utilize Augmented Reality (AR) to enable robot data that would typically be ``hidden''

to be visualized by rendering graphic images over the real world using AR-supported devices (e.g., smartphones, tablets, the Microsoft HoloLens, etc.) \citep{frank2017mobile}. However, AR technology requires access to a physical robot and a real-life location. When real-life HRI experiment is not accessible, a VR simulation with a virtual robot offers an alternative way to conduct HRI studies remotely. 

VR is a computer-generated simulation where a person can interact within an artificial 3D environment. In VR based HRI, a virtual robot and graphic images can resemble the presence in a real-life location \citep{fang2014novel} 
 
There are several advantages to using a virtual robot. A virtual robot can utilize advanced algorithms to behave autonomously like a real robot. For example, studies have compared the interactive behaviors of robots in a VR environment and a real environment. The results confirm that the behaviors of a virtual robot and a robot in a real environment are the same \citep{mizuchi2017cloud}. Virtual robots can also reduce the cost of experiments. Since multiple participants can log in to the VR environment and interact with the virtual robot at the same time, this can minimize the resources needed to conduct the studies in a real environment \citep{mizuchi2017cloud}. 

In this paper, we introduce HAVEN (HRI-based Augmentation in a Virtual robot Environment using uNity), a VR simulation that enables users to interact with a virtual robot. We use \textit{Unity}\footnote{\url{https://unity.com/}}

, a game development engine, to create an interactive VR environment containing a virtual robot and its sensory data and cognitive output that is portrayed with a different type of graphic visualizations. In a technical demonstration of the system, we show that we can create a virtual robot that can act autonomously with advanced algorithms relying only on Unity. We also demonstrated two popular experimental studies in HRI, such as the hallway passing scenario and the human-robot object retrieval scenario. 


\section{RELATED WORK}

 AR provides computer-generated graphics on a user's view of the real world. In the context of mobile robots, the AR robotics system enables users and robots to communicate through a shared reality space that merges the robot’s visual world with the human's \citep{muhammad_creating_2019}. The system consists of different visualizations that represent sensory and cognitive data that is rendered on the user interface. Other studies utilize AR interface to support human-multi-robot collaboration, where the robots collaborate with both the robot and human teammates \citep{chandan2019negotiation}. 

Virtual robots have been utilized in VR-based HRI studies. For example, previous frameworks have combined both robot control (e.g., ROS) and virtual environment display and interaction (e.g., Unity) \citep{codd2014ros}. VR technology enables the reduction of cost in software development as well as other applications such as cloud-based robotics \citep{mizuchi2017cloud} and teleoperation tasks \citep{codd2014ros}. Other studies use Unity as the primary authoring tool for a functional virtual robot \citep{mattingly2012robot}.

HAVEN is a continuation of these previous works, merging the AR graphic interface and a functional virtual robot in VR environment. Compared to the other systems, HAVEN allows the robot to leverage the AR virtualization to communicate with the users. Using only Unity 3D, HAVEN offers simplicity when it comes to building the virtual robot and the VR environment, allowing non-programmers to create VR simulation environment to fit their needs.


\section{SYSTEM ARCHITECTURE AND DESIGN}

\begin{figure}[t]
    \includegraphics[width=0.5\linewidth]{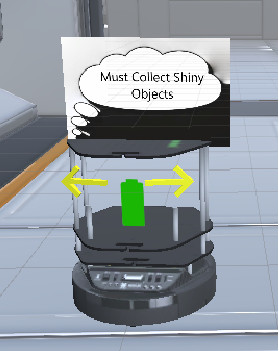}
    \centering

    \caption{Examples of AR visualizations projected by the Turtlebot2 robot. Thought bubble is rendered on top. Battery indicator is in green in the middle, and the yellow arrow are used for motion intent.} 

    \label{fig:turn_signals&thought_bubble_and_battery}
\end{figure}

HAVEN\footnote{The HAVEN repository is available at \url{https://github.com/tufts-ai-robotics-group/HAVEN}} is a modular virtual environment designed to support many different aspects of robotic AR simulation. The functional capabilities of HAVEN include:
\begin{itemize}
    \item Simulate various robot agents of choice. Currently the TurtleBot2 robot is used.
    \item Host a variety of HRI scenarios. HAVEN includes Hallway Passing and Object Collection in single room scenarios. Different environments are in development. 
    \item  Customizable scripts for data visualization and robotic behaviors.
\end{itemize}

\subsection{Software}
Built fully in the \textit{Unity 2018.4} game development engine, all robot AI and other functionality are fully implemented using \textit{Unity's} native C\# script. Researchers who have \textit{Unity} can clone the repository and edit the source code to implement their own algorithms and virtual hardware. The final environment can either be hosted on a cloud service or participants can download a zip that itself contains the executable and all dependencies based on operating system. This flexibility in distribution allows a wide variety of potential participants across several different platforms.   

\subsection{Robot Agent}

The desired robot will be able to be chosen from a list. The given design for these scenarios is a Turtlebot2, currently only equipped with autononmous navigation. The AI is scripted to replicate its motion closely to its real world counterpart.

\subsubsection{Movement Scripting}
The controller script of the TurtleBot2 agent is built on \textit{Unity's} Navigation Mesh (NavMesh) System. NavMesh scans the entire environment and lays a grid atop all navigatable surfaces. NavMeshAgents can then take that grid and mark a destination so long as there exists a valid path and will start to approach it. However, the agent will simultaneously move directly toward the destination and rotate such that to face the destination. To create a more realistic movement, the TurtleBot2 first calculates the signed angle between the direction it is facing and the direction it must face to be looking at the next destination. If the angle's absolute value is above a certain threshold or the destination is too close, then the TurtleBot2 will stop, rotate until it is facing its next destination, then continue forward. Otherwise, the TurtleBot2 will diminish the angle between itself and the destination by traveling in an arc motion until it faces the destination.

\subsubsection{Battery Level}

An optional battery feature, designed to add an extra layer of depth to navigation, is included that is currently designed to decrease in charge relative to translational motion, however different actions . For mobile robots they will always keeps track of how far it is away from its charging base. There are two different modes:
\begin{itemize}
    \item After reaching a destination, before calculating the path to the next destination the TurtleBot2 will calculate the following:
    \begin{itemize}
        \item How far it is from the next destination
        \item How far the next destination is from the base
    \end{itemize}
    If the sum of those calculations exceeds the distance remaining on the battery, the TurtleBot2's next destination becomes the charging base. From there, it will resume reaching the rest of the destinations. 
    \item Every frame the TurtleBot2 compares the distance from the charging base and the remaining distance the battery can travel. As soon as the remaining battery life can travel the distance to the charging base, the TurtleBot2 abandons its current path and reroutes to the charging base. 
\end{itemize}

\subsection{Visualization Options}
HAVEN is currently equipped with 4 different types of visualization:
Path Projection, Turn Signals, Thought Bubbles, and Battery Level Indicator

\subsubsection{Path Projection (Path, Steps to Project)}
Given the TurtleBot2's destination and its rotation requirements, the path is broken up into small segments (the length of the designated markers) and a marker is placed on each segment. The location of each of those markers acts as an intermittent destination when given to NavMesh. The number of markers that appear on-screen at a given time can be specified as a parameter, and there are different markers for a linear and curved motion (See figure \ref{fig:path_projection}). 

\subsubsection{Turn Signals (Angle)}
The sign of the angle calculated between the TurtleBot2 and the destination determines the direction the TurtleBot2 will turn or rotate, and the appropriate direction arrow will appear (See figure \ref{fig:turn_signals&thought_bubble_and_battery}).
\subsubsection{Thought Bubbles (Message string to project)}
When the human agent gets close enough to the TurtleBot2, the message for the TurtleBot2 to display appears above it in a comic style thought bubble that always faces the human agent (See figure \ref{fig:turn_signals&thought_bubble_and_battery}).    
\subsubsection{Battery Level Indicator (Battery Level)}
Similar to the thought bubble, battery level is indicated by a 2D image on on the TurtleBot2 that always faces the human agent (See figure {\ref{fig:turn_signals&thought_bubble_and_battery}}).


\section{APPLICATIONS}

\begin{figure}[t]
    \centering
    \includegraphics[width=1\linewidth]{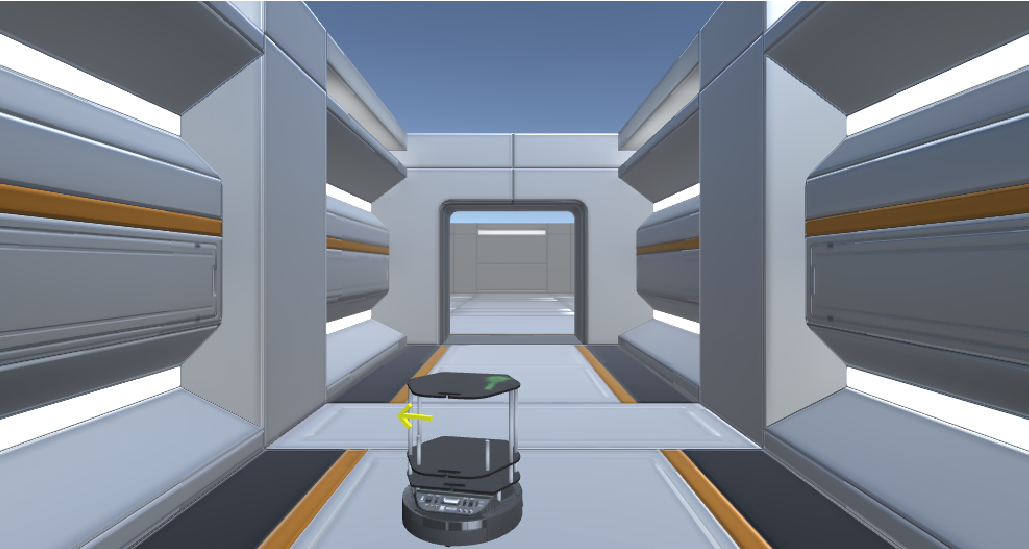}
    \caption{First person view of the hallway passing scenario. The TurtleBot2 detects the human (obstacle) blocking its path and therefore re-plans its path trajectory to the goal. The direction it decides to move around the obstacle is given by the rendered yellow arrow to communicate its motion intent.}
    \label{fig:hallway_scenario}
\end{figure}

HAVEN simulates real-world objectives related to HRI in a format that allows for additions, modifications, and widespread distribution for various questions and research purposes. Scenarios are provided as separate \textit{Unity} scenes built off an initial scenario template. Two example scenarios are included: Hallway traversal and Shared room object retrieval.  

\subsection{Hallway Passing Scenario}

The Hallway Passing Scenario requires both agents to coordinate their movements in order to avoid collision with each other while attempting to reach the other side. HRI researchers have approached human prediction of robot motion intent through external robot modifications, movement behavior scripts, and many other methods. \citep{pacchierotti2005human, watanabe2015communicating, fernandez2018passive}. In HAVEN, the TurtleBot2's motion behavior can be modified using scripts, and it can use visualizations to inform the others of its decisions and intent.
\subsubsection{TurtleBot2's goal}In this scenario, the only visualization utilized is the turn signal. The TurtleBot2's main objective is to reach the center of every room. The TurtleBot2 travels down the center of every hallway unless the human agent is blocking the center path, in which case the TurtleBot2 will start to curve to avoid the human, using the turn signals to indicate its desired direction (see figure \ref{fig:hallway_scenario}).   
\subsubsection{Human's goal}The human's main goal is also to reach the center of every room. In harder scenarios, the human must reach the center of a room before the TurtleBot2 does in order to remove an obstruction that sits in the TurtleBot2's path. The human will always start in a different room as the TurtleBot2 and does not know where the TurtleBot2 starts. 

\subsection{Shared Room Object Retrieval} 

\begin{figure}[h!]
    \centering
    \includegraphics[width=0.9\linewidth]{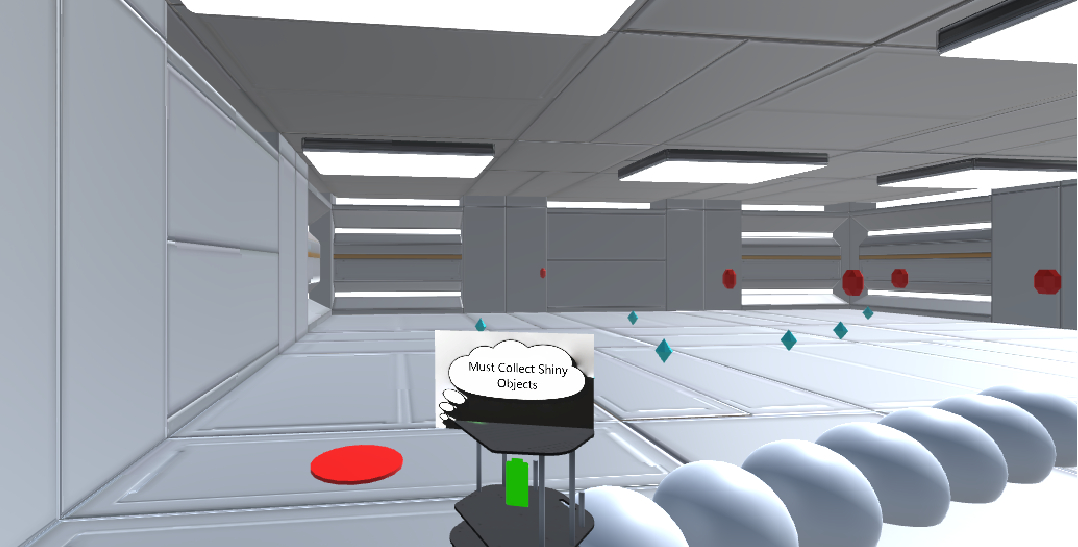}
    \caption{First person view of object retrieval scenario. The red circle on the floor is the TurtleBot2's charging base. The white spheres show the TurtleBot2's path trajectory. Blue gems are robot collectibles. Red gems are human collectibles.}
    \label{fig:object_collection_scenario}
\end{figure}

This scenario was inspired by \citet{walker2018communicating} who conducts a study in which a participant collects beads from a set of stations with a monitor drone. Both agents share the goal of collecting their respective set of gems that are scattered around the room. The TurtleBot2 selects a gem to set its destination to and proceeds to collect it. It will not collect gems it collides with on the way. The human must stand in the gem for a given duration before it is collected. If the human collides with Turtlebot2's path projection it will reroute to curve around the human, unless the human is within a certain distance of either the Turtlebot2 or its destination, in which case it must wait until the human moves away from the path. If the human moves away before the Turtlebot2 reaches the rerouted section it will revert back to its original path (See figure \ref{fig:object_collection_scenario}).

\section{DISCUSSION}

Especially given the circumstances of COIVD-19, there are many benefits to working in simulation over a real-world machine. Robots and visualizations that are currently in the theoretical side can be tested, and studies can be performed. Without being restricted to a lab, participants can access and perform the study from anywhere, either hosted on the web or from downloading the executable and dependencies along with instructions in single zip file. Each study can be fully encapsulated in its own program, given sufficient instructions to participants through text prompts and manuals. The visualizations and scenarios shown in this paper are only the basics of what HAVEN can be capable of. Although not as realistically accurate as most other simulations built on more robotics-oriented software, the versatility of building solely in \textit{Unity} removes many entry barriers for participants and can be reached by a significantly larger audience as either the technical expertise or the proper equipment are required. Not only in terms of participants, but to researchers and developers as well. With \textit{Unity} knowledge and experience, current robots, visualizations, and scenarios can either be modified or built upon to fit specific needs or new ones can be added all together. With more control over the environment, the range of collectible data goes beyond that of robot sensory input, and many environmental variables can be controlled. 


\section{CONCLUSION AND FUTURE WORK}

This paper introduced HAVEN, a VR simulation that enables researchers to conduct AR-based HRI studies to overcome the challenges of social-distancing mandates. With this tool, users can interact with a virtual intelligent robot and see its sensory and cognitive output that is rendered with graphic visualizations projected over a VR world. Then we described two common popular scenarios for experimental AR studies in HRI such as the hallway passing scenarios and the human-robot object retrieval scenarios. 

For future work, we plan to develop a more immersive user experience 
by building HAVEN into head-mounted displays. We also plan on introducing a teleoperation framework for real-time studies that will grant researchers control of the robot, allowing them to interact with participants. In addition, we will consider the types of information that are useful for data collection. We based all our initial work on the TurtleBot2, we plan on extending support for different types of robots as well i.e. drones and manipulators. We expect our system to become a useful tool for HRI researchers to further investigate the effectiveness of proposed algorithms in regard to a robot's movement and how visualizing various robot data can improve human-robot collaboration.

\bibliographystyle{aaai}
\bibliography{references}

\end{document}